\documentclass[10pt,journal,compsoc]{IEEEtran}
\ifCLASSOPTIONcompsoc
  \usepackage[nocompress]{cite}
\else
  \usepackage{cite}
\fi
\ifCLASSINFOpdf
\else
\fi
\usepackage{url}

\usepackage{graphicx}
\usepackage{tikz}
\usepackage{comment}
\usepackage{amsmath,amssymb} 
\usepackage{color}
\usepackage{float}
\usepackage{epsfig}
\usepackage{bm}
\usepackage{multirow}
\usepackage{fancyhdr}
\usepackage{lipsum}
\usepackage{subfigure}
\usepackage{threeparttable}
\usepackage{booktabs}
\usepackage[OT1]{fontenc} 
\usepackage{color}

\begin{document}
%
\title{Hypergraph-based Multi-View Action Recognition using Event Cameras}

%
%

\author{Yue~Gao,~\IEEEmembership{Senior~Member,~IEEE},
        Jiaxuan~Lu,
        Siqi~Li,
        Yipeng~Li,
        Shaoyi~Du,~\IEEEmembership{Member,~IEEE}

\IEEEcompsocitemizethanks{
\IEEEcompsocthanksitem This work was supported by the National Science and Technology Major Project of China under Grant No. 2020AAA0108102. This work was also supported by National Natural Science Funds of China
(No. 62088102, 62021002).
\IEEEcompsocthanksitem Yue Gao and Siqi Li are with BNRist, THUIBCS, BLBCI, School of Software, Tsinghua University, Beijing 100084, China. (gaoyue@tsinghua.edu.cn; lsq19@mails.tsinghua.edu.cn)
\IEEEcompsocthanksitem Jiaxuan Lu is with Shanghai Artificial Intelligence Laboratory, Shanghai 200232, China. (lujiaxuan@pjlab.org.cn)
\IEEEcompsocthanksitem Yipeng Li is with BNRist, THUIBCS, BLBCI, Department of Automation, Tsinghua University, Beijing 100084, China. (liep@tsinghua.edu.cn)
\IEEEcompsocthanksitem Shaoyi Du is with Department of Ultrasound, the Second Affiliated Hospital of Xi'an Jiaotong University. He is also with National Key Laboratory of Human-Machine Hybrid Augmented Intelligence, National Engineering Research Center for Visual Information and Applications, and Institute of Artificial Intelligence and Robotics, Xi'an Jiaotong University, Xi'an 710049, China. (dushaoyi@xjtu.edu.cn)
\IEEEcompsocthanksitem Jiaxuan Lu and Shaoyi Du are the corresponding authors.
}

}

%
%

\markboth{IEEE TRANSACTIONS ON PATTERN ANALYSIS AND MACHINE INTELLIGENCE}%
{Shell \MakeLowercase{\textit{et al.}}: Bare Demo of IEEEtran.cls for Computer Society Journals}
%



\IEEEtitleabstractindextext{%
\begin{abstract} 
Action recognition from video data forms a cornerstone with wide-ranging applications. Single-view action recognition faces limitations due to its reliance on a single viewpoint. In contrast, multi-view approaches capture complementary information from various viewpoints for improved accuracy. 
Recently, event cameras have emerged as innovative bio-inspired sensors, leading to advancements in event-based action recognition.
However, existing works predominantly focus on single-view scenarios, leaving a gap in multi-view event data exploitation, particularly in challenges like information deficit and semantic misalignment.
To bridge this gap, we introduce \textbf{HyperMV}, a multi-view event-based action recognition framework. HyperMV converts discrete event data into frame-like representations and extracts view-related features using a shared convolutional network. By treating segments as vertices and constructing hyperedges using rule-based and KNN-based strategies, a multi-view hypergraph neural network that captures relationships across viewpoint and temporal features is established.
The vertex attention hypergraph propagation is also introduced for enhanced feature fusion.
To prompt research in this area, we present the largest multi-view event-based action dataset $\textbf{THU}^{\textbf{MV-EACT}}\textbf{-50}$, comprising 50 actions from 6 viewpoints, which surpasses existing datasets by over tenfold. Experimental results show that HyperMV significantly outperforms baselines in both cross-subject and cross-view scenarios, and also exceeds the state-of-the-arts in frame-based multi-view action recognition.
\end{abstract}

\begin{IEEEkeywords}
Multi-View Action Recognition, Event Camera, Dynamic Vision Sensor, Hypergraph Neural Network.
\end{IEEEkeywords}}

\maketitle

\IEEEdisplaynontitleabstractindextext

%
\IEEEpeerreviewmaketitle

\IEEEraisesectionheading{\section{Introduction}\label{sec:introduction}}

%
%
%
%

\IEEEPARstart{A}{ction} 
recognition, a fundamental task in computer vision, involves automatically identifying and classifying human actions from video data. It has gained significant attention due to its broad applications in various fields, including surveillance, human-computer interaction, robotics, and video content analysis~\cite{kay2017kinetics,sigurdsson2016hollywood,losey2020controlling,kosch2023survey}. Accurate action recognition enables intelligent systems to understand human behavior, facilitate human-centric applications, and enhance the interaction between humans and machines~\cite{kong2018human}.
With the rapid advancement of multi-camera systems and virtual reality technologies, there is an increasing need to capture and analyze actions from multiple viewpoints. Single-view action recognition methods~\cite{wang2016robust,kay2017kinetics,jiang2019stm,duan2022revisiting} are inherently limited by the viewpoint from which the action is observed. This viewpoint dependency often leads to incomplete understanding and potential misclassification of actions. In contrast, multi-view action recognition approaches~\cite{wang2018dividing,vyas2020multi} offer distinct advantages by integrating information from different viewpoints, which can capture complementary information, leading to more accurate recognition results.

\begin{figure}[t]
  \centering
  \includegraphics[width=0.5\textwidth]{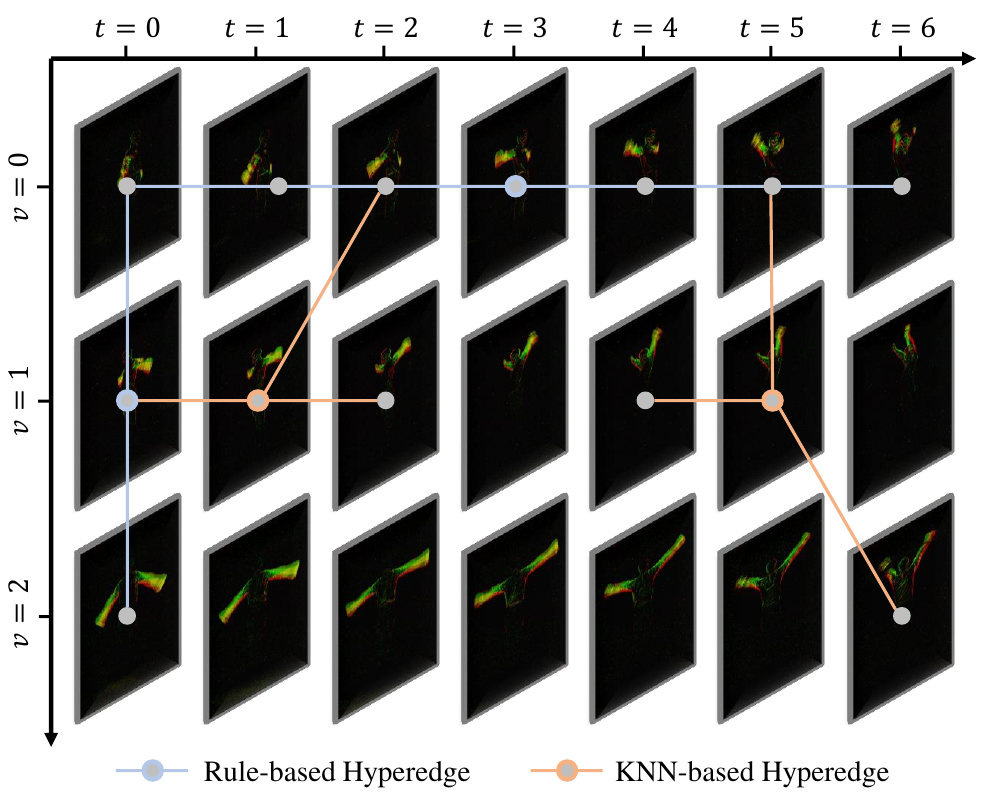}
  \caption{Addressing information deficit and semantic misalignment in multi-view event-based action recognition, the proposed multi-view hypergraph neural network leverages rule-based and KNN-based hyperedges to correlate features across views and temporal segments.}
  \label{Fig:first_page_multi_view}
\end{figure}

Both single-view and multi-view action recognition methods predominantly rely on traditional frame-based cameras and data. As an alternative, bio-inspired event cameras, \textit{e.g.}, Dynamic Vision Sensors (DVS)~\cite{delbruck2016neuromorophic,berner2013240} have emerged as promising vision sensors in recent years. Unlike traditional frame-based cameras that capture images at a fixed exposure rate, event cameras operate by asynchronously detecting changes in brightness at each pixel. This asynchronous event-based feature offers several advantages, including high temporal resolution, low power consumption, and the potential for privacy encryption. While several methods have explored action recognition using event data in single-view settings~\cite{xiao2019event,liu2021event,chen2020dynamic,10198747}, there is currently no existing work that utilizes multi-view event data for action recognition to the best of our knowledge.

One of the reasons is the lack of datasets. Although there are existing single-view event action datasets~\cite{miao2019neuromorphic,calabrese2019dhp19,pradhan2019n,10198747}, there is a lack of comprehensive multi-view event datasets specifically designed for action recognition. DHP19~\cite{calabrese2019dhp19} is the only dataset that can be used for multi-view event-based action recognition, but it is oriented towards pose estimation tasks and is small in scale (33 actions and 2,228 recordings). To facilitate research in multi-view event-based action recognition, we introduce the $\textbf{THU}^{\textbf{MV-EACT}}\textbf{-50}$, an expansion of the single-view $\text{THU}^{\text{E-ACT}}\text{-50}$~\cite{10198747}, by incorporating more viewpoints.
The $\text{THU}^{\text{MV-EACT}}\text{-50}$ dataset comprises 50 distinct actions observed from 6 different views, encompassing 4 frontal and 2 backal views, resulting in a comprehensive collection of 31,500 recording sequences.  The captured dataset stands as the largest multi-view event-based action dataset available to date, which will be released after acceptance.

\begin{figure*}[t]
  \centering
  \includegraphics[width=1.0\linewidth]{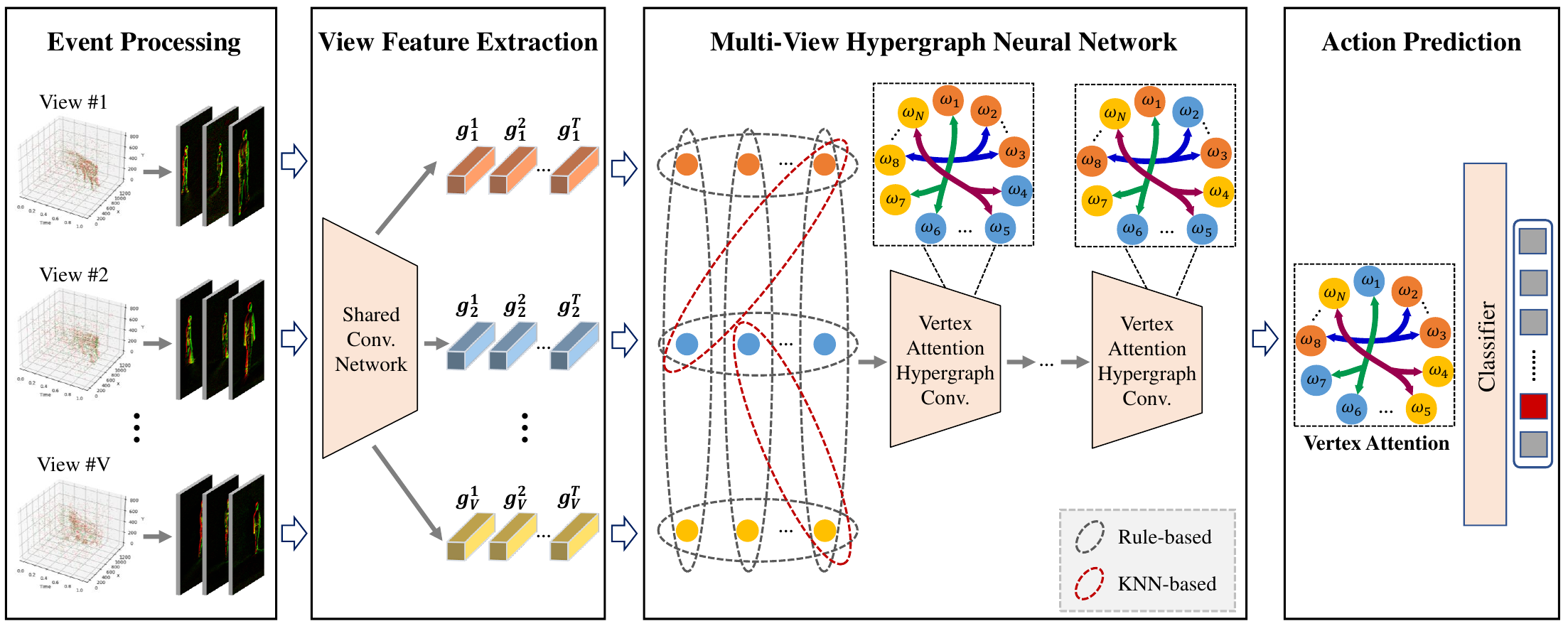}
  \caption{The pipeline of the proposed multi-view event-based action recognition framework, including Event Processing, View Feature Extraction, Multi-View Hypergraph Neural Network, and Action Prediction.}
  \label{Fig:pipeline}
\end{figure*}

In terms of the method, multi-view event data has a serious information deficit compared to frame-based images, since event data only records the regions of motion rather than invariant background. Moreover, event data captured from different viewpoints often encounters a challenge known as semantic misalignment (shown in Figure~\ref{Fig:first_page_multi_view}). Semantic misalignment refers to the inconsistency in pixel position when representing the same region of a person from various viewpoints, a prevalent issue in multi-view scenarios. Therefore, effective fusion of features from diverse viewpoints and moments is pivotal in enhancing the accuracy of multi-view action recognition. 
To address these challenges, we propose a hypergraph-based framework called \textbf{HyperMV} for multi-view event-based action recognition, as shown in Figure~\ref{Fig:pipeline}. The proposed approach explores the high-order associations between viewpoint and temporal features and leverages these associations to facilitate feature fusion. 
Specifically, discrete event data are first processed into frame-like intermediate representations that are fed into the view feature extraction module. A shared convolutional network is acted upon for each view to extract view-related features. In the subsequent stage, each temporal segment under each view is considered as a vertex. By employing both rule-based and K-nearest neighbors (KNN) strategies to construct hyperedges, we establish a multi-view hypergraph neural network to capture both explicit and implicit relationships among viewpoint and temporal features. Vertex attention hypergraph propagation is also proposed for better feature fusion. In the final stage, each vertex is assigned a weight to generate the final embedding, which is subsequently used for action classification.
Extensive experiments involving both cross-subject and cross-view scenarios demonstrate significant improvements compared to the baseline approaches.

Overall, the main contributions of this paper can be summarized as follows:
\begin{itemize}
\item
We extend single-view $\text{THU}^{\text{E-ACT}}\text{-50}$~\cite{10198747} to contribute the $\textbf{THU}^{\textbf{MV-EACT}}\textbf{-50}$ dataset, the largest multi-view event action dataset to date, comprising 50 actions from 6 viewpoints, providing a valuable resource for evaluating algorithms in multi-view event-based action recognition. The constructed benchmarks can be accessed at: \url{https://gaoyue.org/dataset/THU-MV-EACT-50}.
\item 
We propose a framework called \textbf{HyperMV} using the multi-view hypergraph neural network for event-based action recognition, effectively fusing features from different viewpoints and temporal segments.
\item
Through experiments in both cross-subject and cross-view scenarios, we demonstrate the effectiveness of our method with significant improvements in multi-view event-based action recognition compared to the baseline approaches.
\end{itemize}

 

\section{Related Work}
\subsection{Frame-based Action Recognition}
Significant strides have been made in the field of frame-based action recognition~\cite{sun2018optical,wang2015action,lin2019tsm}. Tran \textit{et al.} introduced C3D~\cite{tran2015learning}, a 3D CNN model that merges appearance features with motion data for video sequences. 
Sun \textit{et al.}~\cite{sun2015human} employed factorization techniques to break down 3D convolution kernels and utilized spatio-temporal features across different CNN layers.
The concept of the two-stream CNN~\cite{simonyan2014two} was first introduced to extract features from keyframes and the optical flow channel.  
Wang \textit{et al.} further developed the Temporal Segment Network (TSN)~\cite{wang2016temporal} to utilize video segments within the two-stream CNN framework.
In terms of the multi-branch structure, Feichtenhofer \textit{et al.}~\cite{feichtenhofer2016convolutional} proposed a single CNN that merges spatial and temporal features before the final layers, yielding impressive results. Wang \textit{et al.}~\cite{wang2016two} introduced a multi-branch neural network where each branch handles different levels of features. For the fusion of features at different sample rates, Feichtenhofer \textit{et al.} proposed the SlowFast~\cite{feichtenhofer2019slowfast} which achieves performance improvements by setting fast and slow pathways. In brief, the progression of single-view action recognition is largely dependent on the enhanced aggregation of spatial-temporal features. However, these studies were based on information from a single viewpoint and thus did not learn features from multiple viewpoints.

When it comes to multi-view action recognition, where videos come from various viewpoints, earlier action recognition methods that only utilize view-invariant representations may not deliver optimal results~\cite{turaga2011statistical,zheng2013learning}. Liu \textit{et al.}~\cite{liu2013learning} introduced a genetic algorithm that merges features from various views through a process of iterative evolution. Drawing inspiration from subspace learning, Kong \textit{et al.}~\cite{kong2015bilinear} developed a projection matrix to map features from different views into a shared subspace. In an effort to further progress multi-view learning, Nie \textit{et al.}~\cite{nie2016parameter} endeavored to autonomously learn the optimal weight of each viewpoint without the need for additional parameters. 
Ullah \textit{et al.}~\cite{ullah2021conflux} present a conflux Long Short-Term Memory (LSTM) network to recognize actions from multi-view cameras. For improvement, Bai \textit{et al.}\cite{bai2020collaborative} put forth a collaborative attention mechanism to discern the attention disparities among multi-view inputs. 
Shah \textit{et al.}~\cite{shah2023multi} employ supervised contrast learning to learn feature embedding robust to changes in viewpoint. Another category of methodologies~\cite{Wang_2019_ICCV,liang2022view} made use of Generative Adversarial Networks (GANs) to generate one view conditional on another, thereby probing the potential interconnections between views. Hence, the key to effective multi-view action recognition lies in the fusion of diverse information across various views.

\subsection{Event-based Action Recognition}
Event cameras are bio-inspired vision sensors that asynchronously detect brightness changes at each pixel.
For a given pixel located at $(x,y)$, an event is triggered at a specific timestamp $t$ when the following condition is met:
\begin{equation}
   L(x, y, t)-L(x, y, t') > p \cdot \theta,
\end{equation}
where $L(x, y, t)$ represents the logarithm of the brightness, while $t'$ is the timestamp of the last event that was triggered at the $(x,y)$ location. The constant $\theta$ serves as a threshold, and $p \in \{-1,+1\}$ denotes the polarity of the event, indicating whether the brightness is increasing or decreasing.
Events are recorded as tuples $e=(x_k,y_k,t_k,p_k)$, where $x_k$ and $y_k$ are pixel coordinates, $t_k$ is the timestamp, and $p_k$ is the polarity. 
Compared to traditional frame-based imaging devices, event cameras do not record color or texture, providing a unique privacy advantage. Moreover, they only activate upon significant intensity changes, leading to reduced power usage, about 150 times lower than regular cameras~\cite{walker2011128}. Additionally, in scenarios involving high-speed motion, event cameras can capture motion without the usual blur seen in frame-based cameras.

However, since event cameras are new vision sensors introduced in recent years, only a few works have explored event-based action recognition. Xiao \textit{et al.} introduced the HMAX Spiking Neural Network (HMAX SNN)~\cite{xiao2019event} to extract temporal features via multispike encoding. Building upon this, Liu \textit{et al.} put forth Motion SNN~\cite{liu2021event}, which leverages motion information to construct a multilayer SNN structure. Chen \textit{et al.}~\cite{chen2020dynamic} proposed to view events as 3D points and input them into a dynamic graph CNN for gesture recognition.
To leverage the powerful learning capabilities of CNNs in image-related tasks, several studies have transformed the discrete event data into frame-like representations. Ghosh \textit{et al.}~\cite{ghosh2019spatiotemporal} introduced spatio-temporal filtering in the spike-event domain, with the resulting representations inputted into CNNs. Innocenti \textit{et al.}~\cite{9412991} proposed the conversion of event data into Temporal Binary Representation for subsequent action recognition using CNNs. Wu \textit{et al.}~\cite{wu2020multipath} accumulated the event data into frames and leveraged the multipath deep neural network for action recognition. Gao \textit{et al.}~\cite{10198747} proposed to fuse multiple event representations in a learnable manner and feed them into the event-based slow-fast network for action recognition.
Nonetheless, these methods are all tailored to single-view event-based action recognition, and there is no work exploring multi-view action recognition based on event data to the best of our knowledge.

\subsection{Graph and Hypergraph Neural Network}
In recent years, Graph Neural Networks (GNNs)~\cite{gori2005new,scarselli2008graph} and their variants~\cite{defferrard2016convolutional,duvenaud2015convolutional,ktena2017distance,gao2020hypergraph} have emerged as powerful tools in the realm of data analysis, demonstrating their versatility in a broad spectrum of graph-structured tasks, including graph classification~\cite{zhang2018end,lee2018graph,han2022g}, graph clustering~\cite{tian2014learning,hu2021adaptive,muller2023graph}, and graph link prediction~\cite{cai2020multi,zhu2021neural}. The power of GNNs also extends beyond graph-structured data, as they have also been effectively utilized in non-graph structured data. This includes areas such as document classification~\cite{10.5555/3157382.3157527}, image classification~\cite{quek2011structural,zhou2022few}, person re-identification~\cite{liu2021video,lu2023exploring}, and action recognition~\cite{wang2016graph,yan2018spatial}. Since the GNNs are inherently limited by the graph structure that only allows for one-to-one relationships between vertices, some researchers have turned to hypergraphs~\cite{zhou2006learning} and Hypergraph Neural Networks (HGNNs)~\cite{feng2019hypergraph}. These advanced structures and networks extend the concept of graphs by allowing hyperedges to connect multiple vertices, capable of constructing and learning high-order complex relationships among vertices.
In the context of multi-view event-based action recognition tasks, the features on both the viewpoint and temporal dimensions often suffer from severe information deficit and semantic misalignment. Given these challenges, we posit that GNNs and HGNNs could potentially demonstrate their robust capabilities in association modeling.

\subsection{Datasets for Action Recognition}
Datasets serve as crucial catalysts in advancing deep learning methodologies. In the realm of frame-based action recognition, numerous well-established datasets already exist. The KTH dataset~\cite{schuldt2004recognizing}, an early action dataset, comprises videos of 6 action categories at a resolution of $160\times 120$ across various scenes. The I3DPost dataset~\cite{gkalelis2009i3dpost} offers videos of two individuals interacting, performing 8 different actions. The UCF50 and UCF101 datasets~\cite{soomro2012ucf101} encompass 50 and 101 action categories, respectively, sourced from YouTube. The Kinetics dataset~\cite{kay2017kinetics}, a series of large-scale datasets released by DeepMind, contains 400 action categories with over 400 videos per action.
As for the multi-view action recognition, the NUCLA dataset~\cite{wang2014cross} is captured in UCLA from three different viewpoints, covering 10 action categories performed by 10 subjects. The NTU dataset~\cite{liu2019ntu} stands out with its integration of RGB, depth, and infrared sensors to capture 60 action classes from multiple angles, consisting of 56,880 videos. The PKU-MMD dataset~\cite{liu2017pku} offers a large-scale benchmark for continuous action recognition, containing 1,076 long video sequences in 51 action categories in 3 camera views. The UESTC dataset~\cite{ji2018large} consists of 25,000 sequences across 40 action categories with 8 static viewpoints. The ETRI dataset~\cite{jang2020etri} is a multi-view action recognition dataset for elderly care, which has 112,620 videos captured from 55 action classes across 8 viewpoints. 

Despite the abundance of conventional frame-like datasets, there is a noticeable scarcity of event-based action recognition datasets.
As for the simulated datasets, N-EPIC-Kitchens~\cite{plizzari2022e2} is an event version of the EPIC-Kitchens generated by the event camera simulator. The event UCF-50~\cite{hu2016dvs} is derived from the UCF-50 action recognition dataset, which was captured by displaying its data on a monitor. Regarding the real-world event-based action recognition dataset, PAF~\cite{miao2019neuromorphic} is the first one, which offers 450 recordings spanning 10 categories from an indoor office setting, each with an average length of 5$s$ and a spatial resolution of $346\times 260$.
N-HAR~\cite{pradhan2019n} is another indoor dataset with 3,091 videos, but it is category-unbalanced and contains only 5 actions. 
DailyAction~\cite{liu2021event} provides 1,440 recordings across 12 action categories, albeit with a limited spatial resolution of $128\times 128$ due to acquisition via DVS128~\cite{6407468}. 
\text{$\text{THU}^{\text{E-ACT}}\text{-50}$}~\cite{10198747} vastly expands the scale of data used for single-view action recognition to include 50 action categories and a total of 10,500 recordings.
DHP19~\cite{calabrese2019dhp19} is currently the only dataset available for multi-view event-based action recognition, including 33 sub-actions and a total of 2,228 recordings.  However, DHP19 is primarily designed for the pose estimation task, and the actions are all localized limb movements (\textit{e.g.}, left arm abduction, right arm abduction, left leg knee lift, right leg knee lift) rather than human actions applicable to everyday scenes. Therefore, there exists a pressing demand for large-scale multi-view action recognition datasets captured by event cameras.

\section{Method}
To confront the challenges of information deficit and semantic misalignment, we introduce a hypergraph-based framework for multi-view event-based action recognition, as depicted in Figure~\ref{Fig:pipeline}. Initially, the event processing module transforms discrete event data into frame-like intermediate representations. Afterward, the view feature extraction module extracts view-related features through a shared convolutional network for each viewpoint. Each temporal segment under each view is considered as a vertex, and the multi-view hypergraph neural network based on rule-based and KNN-based strategies is employed to capture both explicit and implicit relationships. The vertex attention mechanism is also utilized in both the proposed vertex attention hypergraph propagation and the final vertex weighting operator, thereby generating the ultimate embedding for action recognition.

\subsection{Event Processing\label{sec:3.2}}
There are two primary strategies in the event data processing. One utilizes Spiking Neural Networks (SNNs) to process event data as impulses~\cite{ghosh2009spiking,xiao2019event,li2021event,9962797}. However, SNNs have limited learning abilities. Alternatively, some methods transform event data into intermediate representations~\cite{benosman2013event,maqueda2018event,zhu2019unsupervised,almatrafi2020distance}, thereby harnessing the advanced learning capabilities of Convolutional Neural Networks (CNNs). In our approach, we follow the latter method and transform the raw event data into the widely used Event Frame~\cite{rebecq2017real}. For a given view $v$, the stream of events $E_{v}$ is decomposed into a sequence of $T$ event packets in temporal order, denoted as $E_{v}=\{E_{v}^{t}\}_{t=1}^T$. Each event packet $E_{v}^{t}$ represents the set of events collected within the time interval from $t-1$ to $t$, represented as 
\begin{equation}
    E_{v}^{t}=\{(x_k,y_k,t_k,p_k)\}_{k=1}^N,
\end{equation}
where $N$ is the total number of events within the time interval from $t-1$ to $t$.

Subsequently, we generate the event frame $I_{v}^{t}$ from the event packet $E_{v}^{t}$ by summing the events triggered at each pixel location for the two polarities, denoted as 
\begin{equation}
    I_{v}^{t} (x,y)=\sum_{e \in E_{v}^{t}} p_k \cdot \delta(x-x_{k}, y-y_{k}),
\end{equation}
where $\delta(\cdot)$ denotes the Dirac delta function, which equals 1 when $x=x_k$ and $y=y_k$, and 0 otherwise. As a result, for each view, the raw event data is transformed into a frame-like intermediate representation $I_{v}=\{I_{v}^{1},I_{v}^{2},\ldots,I_{v}^{T}\}$ with dimensions $(X,Y,T)$. The event frame is straightforward yet effective, as it encapsulates both spatial and temporal information, which is crucial for action recognition. 

\begin{figure}[t]
  \centering
  \includegraphics[width=0.5\textwidth]{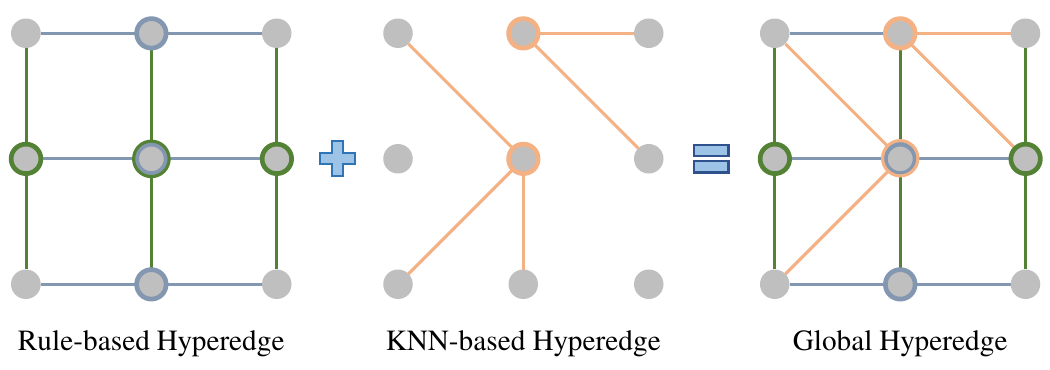}
  \caption{Rule-based and KNN-based hyperedges are combined to model explicit and implicit associations of features.}
  \label{Fig:hypergraph_elements}
\end{figure}

\begin{figure*}[t]
  \centering
  \includegraphics[width=1.0\textwidth]{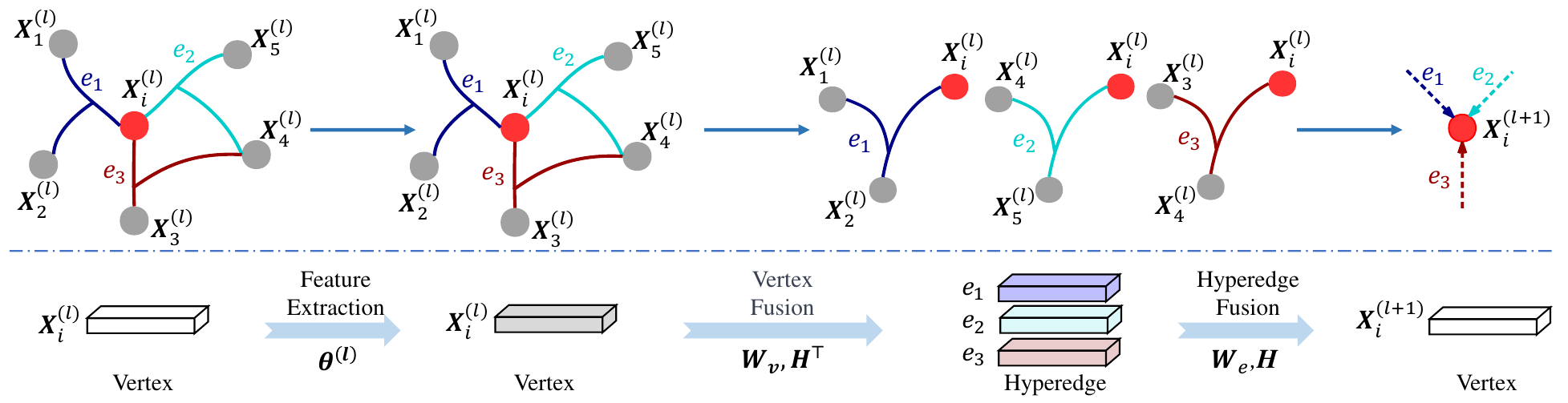}
  \caption{Vertex attention hypergraph propagation via the vertex-hyperedge-vertex process: (1) Initial vertex features enter a fully connected layer for feature extraction. (2) Hyperedge features are then weighted from these vertex features. (3) Subsequently, these hyperedge features are fused together to formulate the vertex features for the next layer.}
  \label{Fig:hypergraph_propagation}
\end{figure*}

\subsection{View Feature Extraction\label{sec:3.3}}
As for the view feature extraction module, our objective is to obtain a comprehensive set of features from different viewpoints. The module operates on input intermediate representations of events, denoted as $I=\left (I_{1}, I_{2}, \ldots, I_{V}\right)$. Each $I_{v}$ corresponds to the intermediate representation from viewpoint $v$. Afterwards, $I$ is processed by a shared convolutional network, which consists of a series of convolutional layers acting as a backbone and a global pooling layer. The shared convolutional backbone is designed to reduce the spatial resolution of each viewpoint representation, thus effectively concentrating vital view-related information. The output can be represented as $C=\left (C_{1}, C_{2}, \ldots, C_{V}\right)$, where $C_{v}=\left\{C_v^t\right\}_{t=1}^T$ contains the feature maps from viewpoint $v$. For each view $v$ and moment $t$, the global pooling layer is then applied to the convolved feature map $c_{v}^t$ to obtain a one-dimensional embedding, denoted as $g_{v}^{t}$.

To formalize, for a given viewpoint $v$ and a moment $t$, the one-dimensional embedding is given by
\begin{equation}
g_{v}^{t} = \text{Pool}\left(\text{Convs}\left( I_{v}^{t} \right) \right),
\end{equation}
where $\text{Convs}(\cdot)$ represents the shared convolutional backbone applied to the intermediate representation $I_{v}^{t}$, and $\text{Pool}(\cdot)$ represents the global pooling layer. The features $g_{v}^{t}$ encapsulate the view-related features of each viewpoint for the moment $t$, which provides a representative embedding for further aggregation at subsequent stages.

\subsection{Multi-View Hypergraph Construction\label{sec:3.4}}
Considering the challenges posed by information deficit and semantic misalignment inherent in multi-view event-based action recognition, the strategy employed for fusing features from various viewpoints and across different temporal segments can greatly influence performance. In a multi-view scenario, there exist both sequential associations across different moments within a single view, and correlations between different views at the same instant. Compared with a normal graph that can only model one-to-one associations, a hypergraph extends the structure of the graph so that multiple vertices can be connected using a single hyperedge. Accordingly, we propose the multi-view hypergraph neural network to integrate features across viewpoints and temporal segments, \textit{i.e.}, using rule-based hyperedges to establish explicit connections, and using KNN-based hyperedges to model implicit connections, as illustrated in Figure~\ref{Fig:hypergraph_elements}.

Specifically, we regard the one-dimensional features $g_{v}^{t}$ under the view $v$ and moment $t$ as the vertices, denoted as $(v,t)$. In the multi-view scenario with $V$ viewpoints and $T$ time windows, there are a total of $V\times T$ vertices. For the rule-based strategy, we employ two types of hyperedges: the time-consistent hyperedge $\mathcal{E}_{rule}^{(t)}$ connects vertices of different moments of the same view, and the view-consistent hyperedge $\mathcal{E}_{rule}^{(v)}$ links vertices from varying views at an identical moment, denotes as
\begin{align}
  \mathcal{E}_{rule}^{(t)}&=\left\{(v,t), \forall (v',t') \mid v=v', t\neq t' \right\}, \\
  \mathcal{E}_{rule}^{(v)}&=\left\{(v,t), \forall (v',t') \mid t=t', v\neq v' \right\}.
\end{align}
Then, the rule-based hyperedges can be denoted as $\mathcal{E}_{rule}=\mathcal{E}_{rule}^{(t)} \cup \mathcal{E}_{rule}^{(v)}$. 
In terms of the KNN-based strategy, we identify for each vertex $(v,t)$ the $k$ vertices in the embedding that exhibit the highest similarity, without consideration for perspective or temporal ordering. These identified vertices are then connected using a hyperedge. As such, the KNN-based hyperedge set $\mathcal{E}_{knn}$ is denoted as
\begin{equation}
  \mathcal{E}_{knn}=\left\{(v,t), \forall (v',t) \in N_k\left(v,t\right) \right\},
\end{equation}
where $N_k\left(v,t\right)$ signifies the $k$ vertices demonstrating the highest similarity to vertex $(v,t)$ in terms of their embeddings. Subsequently, the two types of hyperedge sets are combined to obtain the global hyperedge set $\mathcal{E}=\mathcal{E}_{rule} \cup \mathcal{E}_{knn}$. 
Unlike graphs, hypergraphs utilize the incidence matrix $\textbf{H}$ to indicate whether the hyperedges $e \in \mathcal{E}$ contain the vertices $(v,t)$, which can be expressed as
\begin{equation}
  \textbf{H}((v,t),e) = \begin{cases}1, & (v,t) \in e
  \\ 0, & (v,t) \notin e \end{cases}.
\end{equation}

\begin{table*}[!t]
  \renewcommand\arraystretch{1.3}
  \begin{center}
  \caption{Comparisons between the $\text{THU}^{\text{MV-EACT}}\text{-50}$ benchmark and other existing datasets for single-view and multi-view event-based action recognition.}
  \label{tab:dataset_comparison}
  \setlength{\tabcolsep}{2.5mm}{
  \begin{tabular}{llcccccc} \toprule
  Type & Benchmark & View Num.                        & Category Num. & Recording Num. & Subject Num. & Resolution & Sensor \\ \midrule
  \multirow{4}{*}{Single-view} & PAF~\cite{miao2019neuromorphic} & -   & 10            & 450            & 15             & $346\times 260$   & DAVIS 346      \\
   & DailyAction~\cite{liu2021event} & -  & 12            & 1,440            & 15             & $128\times 128$  & DVS 128  \\ 
   & N-HAR~\cite{pradhan2019n} & -  & 5            & 3,091            & 30             & $304\times 240$  & ATIS  \\ 
   & \text{$\text{THU}^{\text{E-ACT}}\text{-50}$}~\cite{10198747} & -  & 50            & 10,500            & 105             & $1280\times 800$  & CeleX-V  \\ \midrule
  \multirow{2}{*}{Multi-view} & DHP19~\cite{calabrese2019dhp19} & 4  & 33            & 2,228            & 17            & $346\times 260$    & DAVIS 346 \\ 
   & \textbf{$\textbf{THU}^{\textbf{MV-EACT}}\textbf{-50}$} & \textbf{6}            & \textbf{50}   & \textbf{31,500} & \textbf{105}     & \textbf{$1280\times 800$}  & CeleX-V  \\   \bottomrule   
  \end{tabular}}
  \end{center}
  \end{table*}

\subsection{Vertex Attention Hypergraph Propagation\label{sec:3.5}}
After the multi-view hypergraph is constructed, the features of the vertices are updated iteratively based on the connectivity of the hyperedges. 
While the foundational work on Hypergraph Neural Networks (HGNNs)~\cite{feng2019hypergraph} provides a formula for feature propagation through the hypergraph convolutional layer, it solely accounts for the weights associated with the hyperedges, disregarding the weights assigned to the vertices. In our perspective, vertices across distinct viewpoints and moments should possess diverse amounts of information, particularly in the context of event data. With this goal in mind, we propose the vertex attention hypergraph propagation based on the original one, which can be mathematically expressed as
\begin{equation}
\textbf{X}^{(l+1)}=\sigma\left(\textbf{D}_v^{-\frac{1}{2}} \textbf{H} \textbf{W}_e \textbf{D}_e^{-1} \textbf{H}^\top \textbf{W}_v \textbf{D}_v^{-\frac{1}{2}} \textbf{X}^{(l)} \Theta^{(l)}\right),
\end{equation}
where $\textbf{X}^{(l)}$ corresponds to the vertex features of the $l^{th}$ layer. The diagonal matrices, $\textbf{D}_v$ and $\textbf{D}_e$, are formulated from the degree of the vertex and the degree of the hyperedges respectively, and are computed based on the correlation matrix $\textbf{H}$. The nonlinear activation function is represented as $\sigma(\cdot)$. The trainable parameters include $\textbf{W}_e$, $\textbf{W}_v$, and $\Theta^{(l)}$. Here, $\textbf{W}_e$ functions as the weight matrix corresponding to each hyperedge, $\textbf{W}_v$ represents the weight matrix associated with each vertex, while $\Theta^{(l)}$ denotes the weight matrix utilized for feature extraction at the $l^{th}$ layer.

To delve deeper into the feature propagation mechanisms inherent in the vertex attention hypergraph convolution layer, we refer to Figure~\ref{Fig:hypergraph_propagation} which illustrates the vertex-hyperedge-vertex feature fusion process. Initially, the vertex features $\textbf{X}^{(0)}$ are passed through the first fully connected layer with weight $\Theta^{(0)}$ applied to extract relevant features. Subsequently, these extracted features are weighted by $\textbf{W}_v$ to generate each hyperedge's features, represented as the product of the vertex feature matrix $\textbf{X}^{(0)}$ and the transpose of the incidence matrix $\textbf{H}^\top$. Subsequently, the hyperedge features are weighted by $\textbf{W}_e$ to produce the updated vertex features for the next layer $\textbf{X}^{(1)}$, encapsulated by the product of the hyperedge weights $\textbf{W}_e$ and the incidence matrix $\textbf{H}$. Throughout the propagation process, there exists a dynamic interplay between vertex and hyperedge features, which amplifies the capacity to capture intricate high-order relationships. Finally, upon completion of $L$ rounds of hypergraph convolution, we obtain the final vertex features $\textbf{X}^{(L)}$.

\subsection{Action Prediction\label{sec:3.6}}
In the pursuit of effective action classification, it is crucial to merge the final features of $V\times T$ vertices. To this end, we assign different weights to vertices for a superior graph-level representation. Specifically, assuming that the vertex features obtained after the hypergraph convolution layers are represented as $\textbf{X}^{(L)}=\{\textbf{x}_1, \textbf{x}_2, \dots, \textbf{x}_{N}\}$, where $N=V\times T$, the vertex weighting operator computes the attention weight $\omega_{i}$ for each individual vertex. Consequently, the features of each vertex are weighted and amalgamated with their corresponding attention weight, leading to a graph-level feature representation $\textbf{x}_{g}$, denoted as
\begin{equation}
\textbf{x}_{g} = \sum_{i=1}^N \omega_{i} \textbf{x}_i,
\end{equation}
where
\begin{equation}
\omega_{i} = \frac{\left\| \textbf{x}_i \right\|_1}{\sum_j^N \left\| \textbf{x}_j \right\|_1}.
\end{equation}
Finally, the resulting graph-level feature representations $\textbf{x}_{g}$ are fed into the fully connected layer for the prediction of action categories. The cross-entropy loss function is used for training.

\begin{figure*}[!tb]
    \centering
    \includegraphics[width=1.0\textwidth]{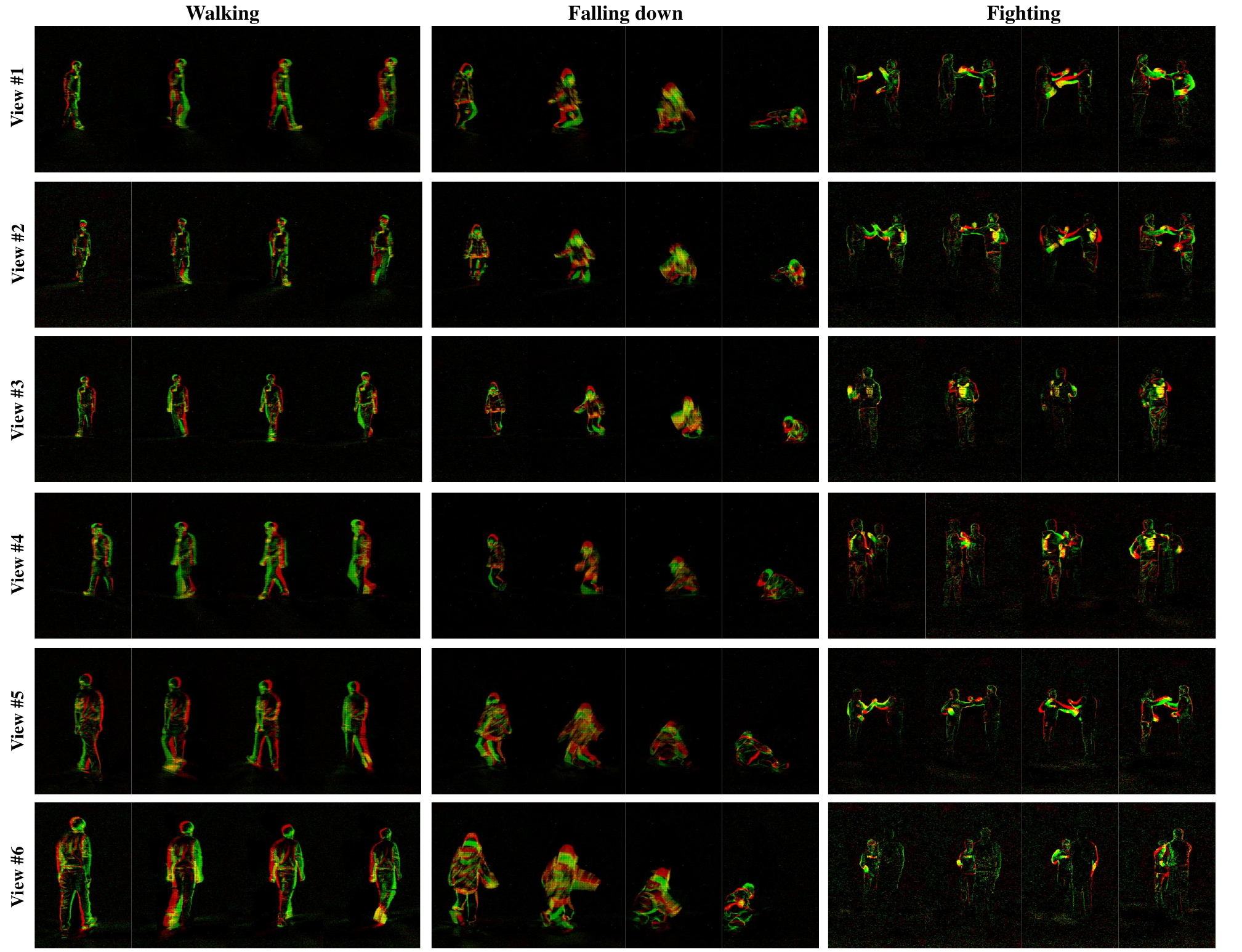}
    \caption{Examples in the collected $\text{THU}^{\text{MV-EACT}}\text{-50}$ dataset.}
    \label{Fig:dataset-v2}
\end{figure*}

\begin{table*}[!tb]
  \renewcommand\arraystretch{1.3}
  \caption{List of actions in the constructed $\text{THU}^{\text{MV-EACT}}\text{-50}$ benchmark.}
  \label{tab:action_category}
  \begin{center}
  \setlength{\tabcolsep}{2mm}{
  \begin{tabular}{lllll} \toprule
  A0: Walking & A10: Cross arms & A20: Calling with phone & A30: Fan & A40: Check time  \\
  A1: Running & A11: Salute & A21: Reading & A31: Open umbrella & A41: Drink water \\
  A2: Jump up & A12: Squat down & A22: Tai chi & A32: Close umbrella & A42: Wipe face\\
  A3: Running in circles & A13: Sit down & A23: Swing objects & A33: Put on glasses & A43: Long jump \\
  A4: Falling down & A14: Stand up & A24: Throw & A34: Take off glasses & A44: Push up \\
  A5: Waving one hand & A15: Sit and stand & A25: Staggering & A35: Pick up & A45: Sit up \\
  A6: Waving two hands & A16: Knead face  & A26: Headache & A36: Put on bag & A46: Shake hands (two-players) \\
  A7: Clap & A17: Nod head & A27: Stomachache & A37: Take off bag & A47: Fighting (two-players) \\
  A8: Rub hands & A18: Shake head  & A28: Back pain & A38: Put object into bag & A48: Handing objects (two-players)\\
  A9: Punch & A19: Thumb up & A29: Vomit & A39: Take object out of bag & A49: Lifting chairs (two-players) \\ \bottomrule         
  \end{tabular}}
  \end{center}
  \end{table*}

\section{$\textbf{THU}^{\textbf{MV-EACT}}\textbf{-50}$ Benchmark}
Regardless of single-view or multi-view scenes, the current event-based datasets are lacking in terms of both action categories and data scale, and contain too simplified actions to meet the data requirements for practical applications of action recognition systems. As shown in Table~\ref{tab:dataset_comparison}, our previously released \text{$\text{THU}^{\text{E-ACT}}\text{-50}$}~\cite{10198747} dataset expands the number of action categories to 50 and recordings to 10,500 for single-view event-based action recognition. However, with respect to multi-view scenarios, DHP19~\cite{calabrese2019dhp19} remains the only applicable choice at present. Although it includes 33 categories of body movements, it focuses on human pose estimation and the simplicity of its limb movement categories restricts its practical applicability.

Given these circumstances, we have extended the single-viewpoint \text{$\text{THU}^{\text{E-ACT}}\text{-50}$}~\cite{10198747} dataset and are about to release the $\text{THU}^{\text{MV-EACT}}\text{-50}$ dataset, which is the first large-scale multi-view dataset specifically for the event-based action recognition task, and also the largest event action dataset to date. 
The $\text{THU}^{\text{MV-EACT}}\text{-50}$ comprises 50 action categories, 31,500 recordings, and 6 viewpoints at a resolution of $1280\times 800$, which surpasses the scale of DHP19~\cite{calabrese2019dhp19} by factors of 14, as shown in Table~\ref{tab:dataset_comparison}.
The $\text{THU}^{\text{MV-EACT}}\text{-50}$ has the same 50 action categories as the \text{$\text{THU}^{\text{E-ACT}}\text{-50}$}, including actions for indoor health monitoring (\textit{e.g.}, falling down, headache, stomachache, back pain, vomit, staggering, etc.), whole-body movements (\textit{e.g.}, walking, running, jump up, running in circles, squat down, tai chi, etc.) and detail-sensitive actions (\textit{e.g.}, nod head, shake head, thumb up, clap, rub hands, wipe face, etc.). At the same time, some confusing action groups are added to increase the difficulty (\textit{e.g.}, stand up, sit down, sit and then stand, etc.). In addition to single-person actions, the dataset also includes actions for interactions between people and objects (\textit{e.g.}, calling with phone, swinging objects, throw, pick up, drink water, open/close umbrella, put on/take off glasses, put on/take off bag, etc.) and actions for interactions between two people (\textit{e.g.}, shake hands, fighting, handing objects, lifting chairs). The complete list of actions is shown in Table ~\ref{tab:action_category}. The props used in the acquisition process include books, umbrellas, school bags, fans, glasses, cups, and tissues.

\begin{figure}[!tb]
    \centering
    \includegraphics[width=0.5\textwidth]{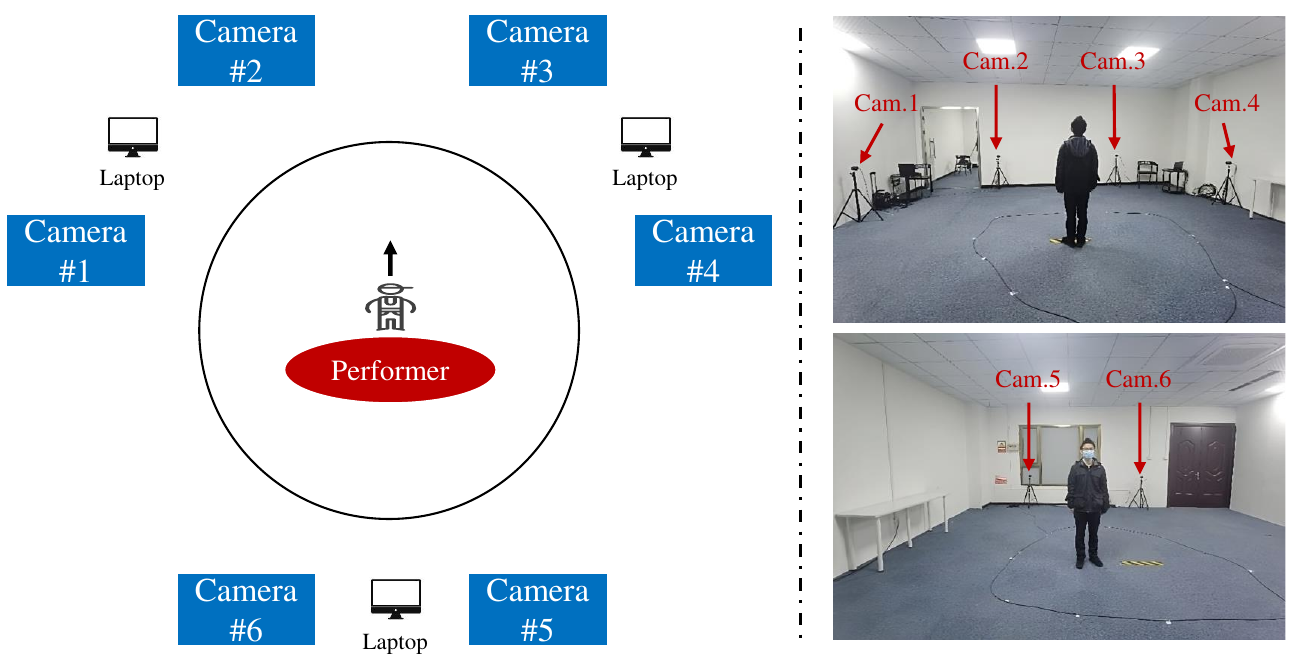}
    \caption{Data acquisition environment of the collected $\text{THU}^{\text{MV-EACT}}\text{-50}$ dataset.}
    \label{Fig:dataset-v2-env}
\end{figure}

For the acquisition environment, the $\text{THU}^{\text{MV-EACT}}\text{-50}$ dataset is collected using CeleX-V~\cite{chen2019live}, through 6 event cameras with different viewpoints arranged across an indoor venue of approximately 100$m^2$. The event cameras, each held by a tripod approximately 1$m$ above the ground, afford 4 frontal and 2 backward views of the performer, as shown in Figure ~\ref{Fig:dataset-v2-env}. Each event camera is adjusted to a fixed orientation to ensure that the performer can be centred in each view of the camera.
Compared to the previous \text{$\text{THU}^{\text{E-ACT}}\text{-50}$} which only contains 2 full front viewpoints (\textit{e.g.}, Camera \#2 and Camera \#3), the $\text{THU}^{\text{MV-EACT}}\text{-50}$ includes additional 2 side-frontal viewpoints and 2 backward viewpoints, making this dataset suitable for multi-view action recognition.
Due to data transmission bandwidth limitations, every two event cameras are connected to a laptop for acquisition. Synchronous triggers ensure simultaneous start and stop of recording across all 6 cameras. The $\text{THU}^{\text{MV-EACT}}\text{-50}$ dataset contains 105 socially recruited subjects, covering all age groups of males and females (15-72 years), which is consistent with the previous \text{$\text{THU}^{\text{E-ACT}}\text{-50}$}. Figure~\ref{Fig:dataset-v2} displays some sampled sequences from all viewpoints. The average duration of each recording across all action categories is 2.34 seconds. Compared with the existing dataset, the $\text{THU}^{\text{MV-EACT}}\text{-50}$ dataset has a total of 31,500 video recordings in 6 perspectives, aiming to provide data support for academic research and application of the multi-view event-based action recognition task.

\section{Experiments and Results\label{sec:5}}
\subsection{Experimental Setup}
Since only the DHP19~\cite{calabrese2019dhp19} and the collected $\text{THU}^{\text{MV-EACT}}\text{-50}$ datasets contain multi-view data, experiments are conducted on these two datasets to verify the effectiveness of the proposed method. In both the DHP19 and $\text{THU}^{\text{MV-EACT}}\text{-50}$ datasets, Top-1, Top-3, and Top-5 accuracy are used as evaluation metrics. In this paper, we have performed experiments in two settings: 1) cross-subject setting, where all viewpoints are input simultaneously during training, allowing us to evaluate the performance of the proposed method using disjoint training and validation or test sets from different performers; and 2) cross-view setting, where the training and test sets are divided based on the viewpoint numbers to assess the generalization ability of the proposed method to unseen viewpoints.

For the cross-subject experiments, we divide the training set, validation set and test set in the ratio of 8:1:1. Specifically, the DHP19 dataset is divided into 12 training objects, 2 validation objects, and 3 test objects, while the $\text{THU}^{\text{MV-EACT}}\text{-50}$ dataset has 85 subjects for training, 10 subjects for validation, and 10 subjects for testing. Since both datasets contain multiple views, they serve as suitable benchmarks for evaluating multi-view action recognition using event cameras.
Regarding the cross-view experiments, we adopt a distinct approach. For the $\text{THU}^{\text{MV-EACT}}\text{-50}$ dataset, 4 views are used for training, 1 view for validation and 1 view for testing. As for the DHP19 dataset, 3 views are employed for training, and 1 view is used for testing. Notably, both the training and test sets encompass all subjects, ensuring comprehensive coverage.

\begin{table*}[!tb]
\renewcommand\arraystretch{1.4}
\begin{center}
\caption{Comparative analysis of the accuracy of single-view, multi-view baselines, and the proposed method across cross-subject and cross-view scenes.}
\label{tab:compare_baseline}
\setlength{\tabcolsep}{2.0mm}{
\begin{tabular}{l|l|cccccc|cccccc}
\toprule
\multirow{3}{*}{Type}  & \multirow{3}{*}{Method} & \multicolumn{6}{c|}{Cross-subject}                                               & \multicolumn{6}{c}{Cross-view}                                                  \\ \cline{3-14} 
 & & \multicolumn{3}{c|}{DHP19~\cite{calabrese2019dhp19}}                 & \multicolumn{3}{c|}{$\text{THU}^{\text{MV-EACT}}\text{-50}$} & \multicolumn{3}{c|}{DHP19~\cite{calabrese2019dhp19}}                 & \multicolumn{3}{c}{$\text{THU}^{\text{MV-EACT}}\text{-50}$} \\
 & & Top-1 & Top-3 & \multicolumn{1}{c|}{Top-5} & Top-1      & Top-3      & Top-5     & Top-1 & Top-3 & \multicolumn{1}{c|}{Top-5} & Top-1      & Top-3     & Top-5     \\ \midrule
\multirow{2}{*}{Single-view} & Baseline    & 78.94 & 94.26 & \multicolumn{1}{c|}{99.38} & 89.98      & 96.85      & 98.34     & 30.63 & 56.62 & \multicolumn{1}{c|}{70.08} & 47.98      & 66.95     & 77.82     \\
 & EV-ACT~\cite{10198747}   & 82.64 & 95.31 & \multicolumn{1}{c|}{99.58} &  92.15     & 97.53      &  98.74    & 37.25  & 63.23 & \multicolumn{1}{c|}{75.34} &  51.26    & 70.13     & 77.69     \\ \midrule
\multirow{3}{*}{Multi-view}  & Baseline     & 81.43 & 95.20 & \multicolumn{1}{c|}{99.10} & 92.53      & 97.59      & 98.84     & 39.28 & 65.47 & \multicolumn{1}{c|}{76.63} & 54.38      & 73.46     & 80.39     \\ 
 & HyperMV-GNN             & 85.70 & 97.55 & \multicolumn{1}{c|}{99.63} & 94.82      & 98.47      & 99.88     & 47.62 & 66.61 & \multicolumn{1}{c|}{77.95} & 56.43      & 74.26     & 82.13     \\
 & HyperMV            & \textbf{92.42} & \textbf{98.65} & \multicolumn{1}{c|}{\textbf{100}}   & \textbf{95.74}      & \textbf{99.23}      & \textbf{99.90}     & \textbf{51.63} & \textbf{67.18} & \multicolumn{1}{c|}{\textbf{78.49}} & \textbf{58.54}      & \textbf{78.07}     & \textbf{83.92}     \\ \bottomrule
\end{tabular}}
\end{center}
\end{table*}

\subsection{Implementation Details}
The key aspect of the proposed method lies in utilizing hypergraph neural networks to capture high-order associations among features across different viewpoints and temporal segments. To demonstrate the effectiveness of each proposed module, we establish a multi-view baseline method that directly fuses these features after converting the raw event data into a frame-like representation and passing them through the view feature extraction module during training. Notably, the multi-view baseline approach does not apply the hypergraph neural network and the vertex attention mechanism. Moreover, we also investigate the application of graph neural network structure and the direct utilization of a single-view baseline network, both of which are elaborated upon in subsequent subsections.

For the main experiments, ResNet 18~\cite{resnet} pre-trained on ImageNet~\cite{deng2009imagenet} is used as the backbone network to meet the low-power requirements of event processing. The number of time windows $T$ in event processing is set to 9, the number of $k$ in the KNN-based hyperedge is set to 3, and the number of $L$ in hypergraph propagation is set to 2. The network has been trained for 40 epochs on all benchmarks using the Adam~\cite{adam} optimizer with an initial learning rate value of $1\times 10^{-4}$, a weight decay of $1\times 10^{-4}$ and a batch size of 12. The exponential learning rate decay~\cite{li2019exponential} strategy is applied, with a gamma of 0.5. All experiments are implemented based on PyTorch~\cite{paszke2017automatic}, and training with a Tesla V100 GPU. 

\begin{table}[!tb]
\renewcommand\arraystretch{1.3}
\begin{center}
\caption{Comparison with SOTA of frame-based multi-view action recognition in terms of Top-1 accuracy.}
\label{tab:compare_sota}
\setlength{\tabcolsep}{0.84mm}{
\begin{tabular}{l|cc|cc}
\toprule
\multirow{2}{*}{Method}                                         & \multicolumn{2}{c|}{Cross-subject} & \multicolumn{2}{c}{Cross-view} \\ 
    & DHP19       & $\text{THU}^{\text{MV-EACT}}\text{-50}$       & DHP19     & $\text{THU}^{\text{MV-EACT}}\text{-50}$     \\ \midrule
CNN-BiLSTM~\cite{zhu2019unsupervised}   & 73.43       & 84.34                & 30.05     & 39.90              \\
Att-LSTM~\cite{zhang2018adding}   & 76.03       & 86.32                & 34.10     & 41.29              \\
DA-NET~\cite{wang2018dividing}   & 84.07       & 92.85                & 43.58     & 51.10              \\
CVAction~\cite{vyas2020multi}   & 85.46       & 93.52                & 46.42     & 54.26              \\
ViewCLR~\cite{das2023viewclr}   & 90.28       & 94.21                & 47.78     & 55.82              \\ \midrule
HyperMV & \textbf{92.42}       & \textbf{95.74}                & \textbf{51.63}     & \textbf{58.54}              \\ \bottomrule
\end{tabular}}
\end{center}
\end{table}

\subsection{Quantitative Results}
\subsubsection{Cross-subject Evaluation}
In the cross-subject experiments, we divide the training, validation and test set based on the number of subjects, and simultaneously input data from all views of a given sample for action classification in the training phase. We set up two baselines for comparison: a single-view and a multi-view, both of which transform the raw event data into frame-like representations. In the training phase, the single-view baseline only inputs one view at a time, while the multi-view baseline employs the view feature extraction module to input multiple views and concatenates the features obtained from each view for action classification. Further, we also construct a GNN-based method named "HyperMV-GNN" on top of the multi-view baseline. For the rule-based strategy, it connects features from adjacent time sequences within the same view and connects features from different views within the same time. In the knn-based strategy, each vertex is connected to its $k$ most similar vertices via $k$ vertex-to-vertex edges. The graph convolution operation and the vertex attention mechanism are utilized to fuse features. To encapsulate our entire model, the proposed complete framework based on Hypergraph Neural Network (HGNN) is referred to as "HyperMV".

Table~\ref{tab:compare_baseline} presents the recognition accuracies of various methods on the DHP19 and $\text{THU}^{\text{MV-EACT}}\text{-50}$ datasets under the cross-subject setting. In addition to the single-view baseline, we also employ EV-ACT~\cite{10198747}, the current SOTA method of single-view action recognition for comparison.
When compared to the single-view baseline, the multi-view baseline improves Top-1 accuracy by 2.49\% and 2.55\% on the DHP19 and $\text{THU}^{\text{MV-EACT}}\text{-50}$ datasets, respectively. Even comparing to the single-view SOTA method, the multi-view baseline can approach or even exceed the accuracy of EV-ACT~\cite{10198747}, demonstrating the advantages of using information from multiple viewpoints. Furthermore, the GNN and HGNN-based methods achieve higher accuracy compared to the multi-view baseline, thanks to their superior fusion of features from different viewpoints and temporal sequences. Specifically, HyperMV-GNN enhances Top-1 accuracy by 4.27\% and 2.29\% on the two datasets. Meanwhile, HyperMV can boost Top-1 accuracy by 10.99\% and 3.21\%, respectively. Due to the hypergraph's capability to model high-order correlations (\textit{e.g.}, feature fusion from the same viewpoint at any moment via 1-hop), HyperMV holds an advantage over HyperMV-GNN, particularly on small datasets like DHP19.

\subsubsection{Cross-view Evaluation}
Cross-view evaluation aims to test the model's generalization capacity for unseen views. Specifically, for the DHP19 dataset, we use 3 viewpoints for training and 1 for testing, while for the $\text{THU}^{\text{MV-EACT}}\text{-50}$ dataset, 4 viewpoints are used for training, 1 viewpoint for validation and 1 viewpoint for testing. As demonstrated in Table~\ref{tab:compare_baseline}, all baselines and methods encountered significant accuracy degradation compared to the cross-subject setting. The single-viewpoint baseline suffers a decrease in Top-1 accuracy by 48.3\% and 42.0\% on the DHP19 and $\text{THU}^{\text{MV-EACT}}\text{-50}$ datasets, respectively. Comparatively, the multi-view baseline enhances Top-1 accuracy by 8.65\% and 6.40\% on the two datasets due to its ability to explore feature associations between different views during training. The GNN and HGNN-based methods proposed in our paper demonstrate stronger cross-view generalization capacities. In particular, HyperMV-GNN further improves Top-1 accuracy by 8.34\% and 2.05\% over the multi-view baseline on both datasets. Meanwhile, HyperMV enhances Top-1 accuracy by 12.35\% and 4.16\% compared to the multi-view baseline. The more pronounced performance improvement on the DHP19 dataset compared to the $\text{THU}^{\text{MV-EACT}}\text{-50}$ dataset can primarily be attributed to the fact that DHP19 encompasses only four views. This condition makes the proposed GNN and HGNN-based method more effective in augmenting model generalization across views, particularly when handling complex association modeling.

\begin{table}[!tb]
\renewcommand\arraystretch{1.3}
\begin{center}
\caption{Comparative analysis of Top-1 accuracy across different hypergraph construction strategies.}
\label{tab:compare_knn_rule}
\setlength{\tabcolsep}{1.1mm}{
\begin{tabular}{l|cc|cc}
\toprule
\multirow{2}{*}{Strategy}                                         & \multicolumn{2}{c|}{Cross-subject} & \multicolumn{2}{c}{Cross-view} \\ 
    & DHP19       & $\text{THU}^{\text{MV-EACT}}\text{-50}$       & DHP19     & $\text{THU}^{\text{MV-EACT}}\text{-50}$     \\ \midrule
Rule-based   & 90.42       & 93.89                & 47.14     & 57.18              \\
KNN-based   & 91.03       & 93.15                & 47.80     & 56.92              \\
\begin{tabular}[c]{@{}l@{}}Rule-based\\  + KNN-based\end{tabular} & \textbf{92.42}       & \textbf{95.74}                & \textbf{51.63}     & \textbf{58.54}              \\ \bottomrule
\end{tabular}}
\end{center}
\end{table}

\subsubsection{Comparisons with Frame-based Methods}
Since there exists no work on event-based multi-view action recognition, we compare the proposed HyperMV with several classical works in frame-based multi-view action recognition. Specifically, we view the event frames processed by Event Processing module as natural images, which are then fed into frame-based frameworks, including CNN-BiLSTM~\cite{zhu2019unsupervised}, Att-LSTM~\cite{zhang2018adding}, DA-NET~\cite{wang2018dividing}, CVAction~\cite{vyas2020multi}, and ViewCLR~\cite{das2023viewclr}.
As detailed in Table~\ref{tab:compare_sota}, the results show that HyperMV outperforms the state-of-the-art (SOTA) methods in both cross-subject and cross-view scenarios for both two datasets.
Specifically, on the DHP19 dataset, HyperMV achieves a 2.14\% and 5.21\% increase in Top-1 accuracy over the SOTA's ViewCLR~\cite{das2023viewclr} in cross-subject and cross-view scenarios, respectively. Similarly, on the $\text{THU}^{\text{MV-EACT}}\text{-50}$ dataset, it shows improvements of 1.53\% and 2.72\% in two scenarios, respectively. These results indicate that HyperMV holds significant advantages in multi-view action recognition for event data. The reason mainly lies in the proposed multi-view hypergraph neural network, which effectively integrates features across various viewpoints and moments, alleviating the critical issues of information deficit and semantic misalignment often encountered in event-based multi-view action recognition.

\begin{figure}[!tb]
  \begin{minipage}{0.245\textwidth}
    \centering
    \includegraphics[width=\linewidth]{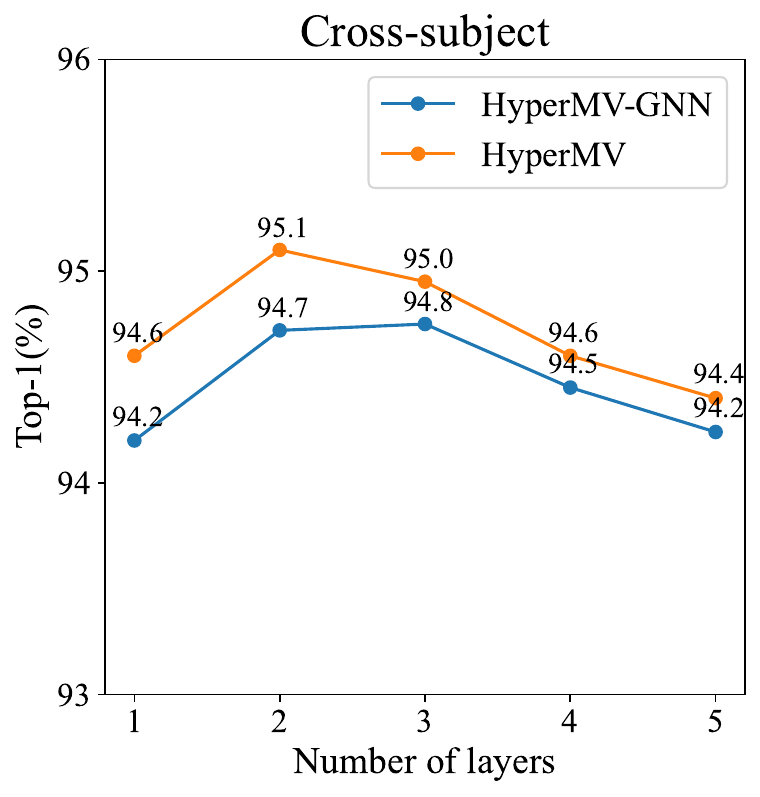} 
  \end{minipage}\hfill
  \begin{minipage}{0.245\textwidth}
    \centering
    \includegraphics[width=\linewidth]{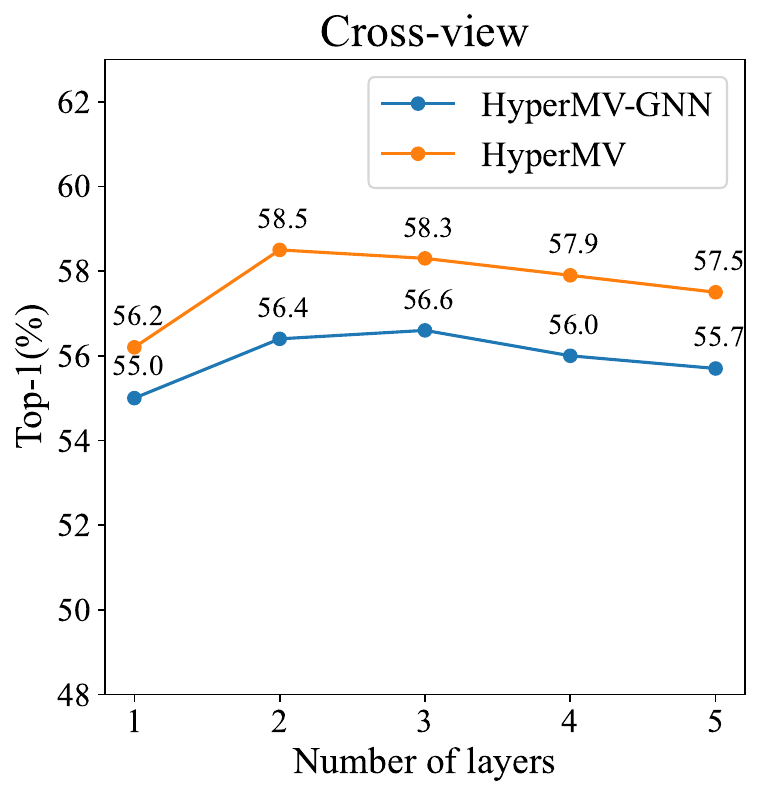} 
  \end{minipage}
  \caption{Impact of varying hypergraph propagation layers on the Top-1 accuracy.}
  \label{Fig:compare_hgnn_layers}
\end{figure}

\subsection{Component Analysis\label{sec:5.4}}
\textbf{Hypergraph Construction Strategy}. 
The proposed hypergraph construction strategy integrates both rule-based and KNN-based hyperedges. To evaluate the effectiveness of these different strategies, we perform experiments in three settings: purely rule-based, purely KNN-based, and a combination of both. All settings are based on the proposed multi-view hypergraph neural network. The results are outlined in Table~\ref{tab:compare_knn_rule}. The KNN-based strategy outperforms the rule-based strategy on the DHP19 dataset in both cross-subject and cross-view scenes. Conversely, the rule-based strategy yields better results on the $\text{THU}^{\text{MV-EACT}}\text{-50}$ dataset. 
However, the highest performance is achieved across all scenarios when both types are employed. Relative to the most effective individual strategy, the simultaneous use of both hyperedges improves the Top-1 accuracy by 1.39\% and 1.85\% on the DHP19 and $\text{THU}^{\text{MV-EACT}}\text{-50}$ datasets in the cross-subject setting, and by 3.83\% and 1.36\% in the cross-view setting. These findings suggest that explicit and implicit associations between perspectives and temporal sequences offer distinct benefits on different datasets. Simultaneously leveraging both strategies enables the system to draw from rule-based correlations as well as uncover implicit associations.

\textbf{Number of Hypergraph Layers}.
To investigate the impact of the number of hypergraph convolutional layers on multi-view event-based action recognition performance, we conduct experiments using two network structures, ~\textit{i.e.}, HyperMV-GNN and HyperMV, with varying numbers of layers. For the $\text{THU}^{\text{MV-EACT}}$ dataset, we perform feature fusion across views and temporal segments under both cross-subject and cross-view scenes, with the number of layers $L$ ranging from 1 to 5. The results are illustrated in Figure~\ref{Fig:compare_hgnn_layers}. According to the experimental outcomes, HyperMV consistently outperforms the HyperMV-GNN approach, which suggests that HGNN possess superior capabilities in establishing feature correlation. As for the varying number of layers, alterations in the number of graph and hypergraph layers can influence recognition accuracy. In the HyperMV-GNN, the best Top-1 accuracy in both scenarios is achieved with $L=3$, obtaining 94.8\% under the cross-subject and 56.6\% under the cross-view setting. Given that HyperMV considers more complex correlations, it achieves optimal performance with $L=2$, yielding Top-1 accuracies of 95.1\% and 58.5\% respectively. However, both HyperMV-GNN and HyperMV experience a decrease when the number of layers continues to increase. This might be attributed to the "over-smoothing" phenomenon caused by excessive feature fusion, in which vertex features converge through overly extensive propagation, thereby diminishing the network's ability to capture locally differentiated features. Hence, it is advisable to set the number of graph and hypergraph convolution layers within the range of $L \in [2,3]$.

\begin{table}[!tb]
\renewcommand\arraystretch{1.0}
\begin{center}
\caption{Top-1 accuracy on the $\text{THU}^{\text{MV-EACT}}\text{-50}$ dataset with or without the vertex attention mechanism.}
\label{tab:compare_att}
\setlength{\tabcolsep}{2.0mm}{
\begin{tabular}{l|c|c|c}
\toprule
Method                               & Attention         & Cross-subject & Cross-view \\ \midrule
\multirow{2}{*}{Multi-view Baseline} & $\times$             & 91.35         & 53.19      \\
 & $\checkmark$ & \textbf{92.53}         & \textbf{54.38}      \\ \midrule
\multirow{2}{*}{HyperMV-GNN}         &  $\times$            & 93.02         & 55.12      \\
 & $\checkmark$ & \textbf{94.82}         & \textbf{56.43}      \\ \midrule
\multirow{2}{*}{HyperMV}        & $\times$             & 93.42         & 57.21      \\
 & $\checkmark$ & \textbf{95.74}         & \textbf{58.54}      \\ \bottomrule
\end{tabular}}
\end{center}
\end{table}

\textbf{Vertex Attention Mechanism}.
The vertex attention mechanism in this paper encompasses both the proposed vertex attention hypergraph propagation and the final vertex weighting operator. To validate the efficacy of the vertex attention mechanism, we perform ablation experiments on the $\text{THU}^{\text{MV-EACT}}\text{-50}$ dataset with and without the use of the vertex attention mechanism. In addition to the experiments on HyperMV-GNN and HyperMV, we also test the effects of using the traditional attention weight (\textit{i.e.}, weighting the features obtained from $V$ views) on the multi-view baseline. As shown in Table~\ref{tab:compare_att}, the utilization of the attention mechanism yields an enhancement in Top-1 recognition accuracy in both cross-subject and cross-view settings. The vertex attention mechanism improves by 1.80\% and 1.31\% under HyperMV-GNN for cross-subject and cross-view scenarios, and by 2.32\% and 1.33\% under HyperMV, respectively. Additionally, the conventional attention mechanism also provides an increase in recognition accuracy under the multi-view baseline, improving by 1.18\% and 1.19\% in the two scenarios. These results indicate that there exist discrepancies in the significance of event features under different viewpoints and moments, and assigning attention weights to these features can result in better performance.

\begin{figure}[!tb]
  \begin{minipage}{0.245\textwidth}
    \centering
    \includegraphics[width=\linewidth]{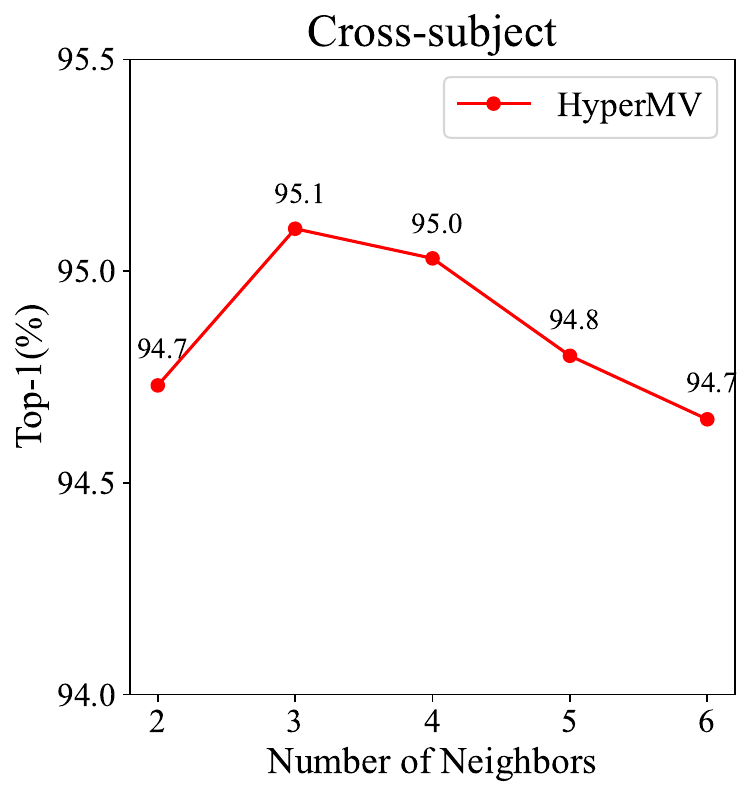} 
  \end{minipage}\hfill
  \begin{minipage}{0.245\textwidth}
    \centering
    \includegraphics[width=\linewidth]{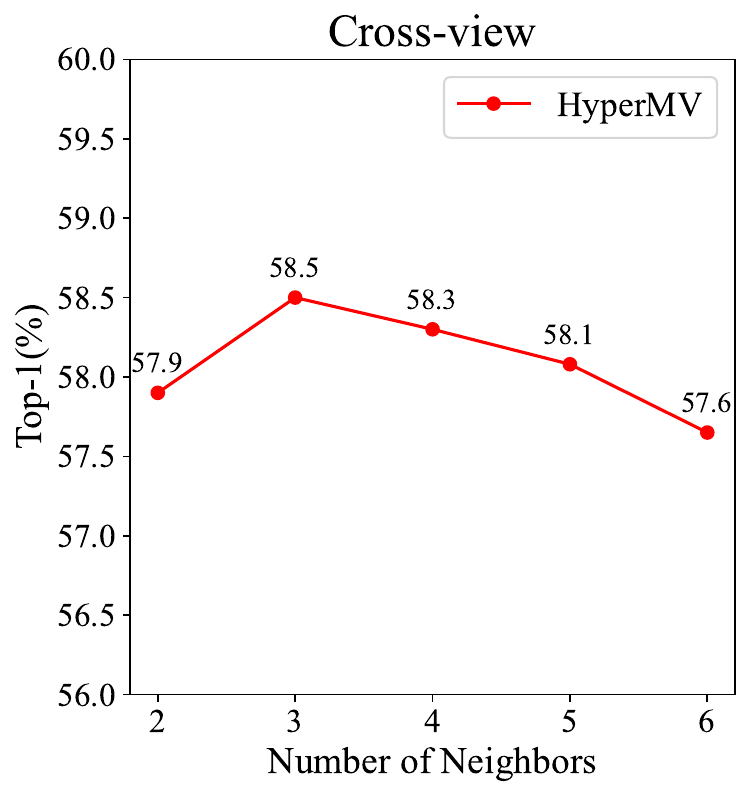} 
  \end{minipage}
  \caption{Impact of varying number of neighbors in the KNN-based hyperedge on the Top-1 accuracy.}
  \label{Fig:compare_knn}
\end{figure}

\textbf{KNN-based Hyperedge}.
The hypergraph construction approach proposed in this paper incorporates a KNN-based hyperedge strategy, which establishes hyperedge connections based on the K-nearest neighbor of vertex embedding. To assess the impact of the number of neighbors on the performance, we examine changes in Top-1 recognition accuracy by modifying the value of $k$ on the $\text{THU}^{\text{MV-EACT}}\text{-50}$ dataset, as shown in Figure ~\ref{Fig:compare_knn}. It can be observed that when $k$ is small (\textit{e.g.}, $k=2$), the limited quantity of neighbors hampers the hypergraph's capacity to model complex relationships among vertices, devolving into a standard GNN. In the majority of cases, a slight increase in $k$ aids the model in capturing distant feature associations better, thereby improving recognition accuracy. For instance, when $k=3$, the proposed method attains the highest Top-1 recognition accuracy in both cross-subject and cross-view settings (95.1\% and 58.5\%, respectively). However, a further increase of $k$ results in a decline in accuracy. For example, when $k=6$, the recognition accuracy drops to 94.7\% and 57.6\% respectively. This suggests that an excessively large number of K-nearest neighbors leads the model to over-fuse information with significant variations in viewpoint and temporal sequence. Consequently, the number of neighbors can have some slight effects on the performance of the multi-view event-based action recognition.

\subsection{Model Complexity\label{sec:5.5}}
To quantitatively assess the efficiency of the proposed HyperMV, we conduct experiments on the model complexity within the cross-subject test set of $\text{THU}^{\text{MV-EACT}}\text{-50}$. The model parameters (M) and the number of floating-point operations per second (GFlops) are employed as evaluation metrics. Three classical multi-view frame-based action recognition methods are selected for comparisons, including DA-NET~\cite{wang2018dividing}, CVAction~\cite{vyas2020multi} and ViewCLR~\cite{ das2023viewclr}. The results shown in Table~\ref{tab:model_complexity} indicate that the proposed framework requires only 22.8M parameters and 31.2 GFlops per viewpoint, substantially undercuts the complexity of all frame-based counterparts. Among them, CVAction emerges as the most complex due to its reliance on a 3D CNN backbone. In contrast, HyperMV achieves better performance in multi-view event-based action recognition with a significantly lower complexity.

\begin{table}[!tb]
\renewcommand\arraystretch{1.3}
\begin{center}
\caption{Comparison of event-based and frame-based multi-view action recognition in terms of model complexity.}
\label{tab:model_complexity}
\setlength{\tabcolsep}{4.5mm}{
\begin{tabular}{l|c|c}
\toprule
Method  & Params (M) & Flops (G) $\times$ Views \\ \toprule
DA-NET~\cite{wang2018dividing}   & 28.5       & 75.6 $\times$ 6     \\
CVAction~\cite{vyas2020multi}   & 72.1       & 183.4 $\times$ 6 \\
ViewCLR~\cite{das2023viewclr}   & 32.5       & 72.4 $\times$ 6 \\ \midrule
HyperMV & \textbf{22.8}       & \textbf{31.2 $\times$ 6}  \\ \bottomrule
\end{tabular}}
\end{center}
\end{table}

\section{Conclusion}
In this paper, we introduce \textbf{HyperMV}, a pioneering framework for multi-view event-based action recognition. The proposed framework converts the inherently discrete event data into frame-like representations for each viewpoint. 
Through the implementation of the multi-view hypergraph neural network by employing rule-based and KNN-based strategies, the framework not only augments the capacity to capture explicit and implicit feature associations but also adds an extra dimension to the current landscape of multi-view action recognition research. Moreover, HyperMV incorporates a vertex attention mechanism to further enhance its action recognition efficacy. 
We also contribute to the broader research community by constructing the largest multi-view event action dataset, \textit{i.e.}, $\textbf{THU}^{\textbf{MV-EACT}}\textbf{-50}$, which will serve as a significant resource for future academic evaluations and real-world applications. 
\ifCLASSOPTIONcaptionsoff
  \newpage
\fi



%




\bibliographystyle{IEEEtran}
\bibliography{egbib}

\end{document}